# UZBEK AFFIX FINITE STATE MACHINE FOR STEMMING

**Sharipov Maksud**
**PhD, Associate Professor**
**Department "Information Technologies"**
**Urgench State University**
**E-mail: maqsbek72@gmail.com**

**Ulugbek Salaev**
**PhD student, Department**
**"Information Technologies"**
**Urgench State University**
**ulugbek0302@gmail.com**
**Yuldashov Ollabergan**
**Master student, Department**
**"Information Technologies"**
**Urgench State University**

**Sobirov Jasur**
**Master student Department**
**"Information Technologies"**
**Urgench State University**

**Annotation:** This work presents morphological analyzer for Uzbek language using finite state machine. The proposed methodology is morphologic analyzing of Uzbek words by using an affix striping to find a root and without including any lexicon. This method helps to perform morphologic analysis of a words from a large amount of text in high speed as well as it is not required using of memory for keeping vocabulary. According Uzbek is an agglutinative language can be designed finite state machines (FSMs). In contrast to the previous works, in this study are modeled the completed FSMs for all word classes by using the Uzbek language's morphotactic rules in right to left order. This paper shows the stages of this methodology including the classification of the affixes, the generation of the FSMs for each affix class and combine into a head machine to make analysis a word.
**Key Words:** Natural Language Processing, Morphology, Derivational, Inflectional, Finite State Machine, Affix Stripping, Uzbek morphology, Stemming.
**Аннотация.** В работе представлен морфологический анализатор узбекского языка. Предлагаемая методика представляет собой анализ узбекских слов с использованием чередования аффиксов для поиска корня и без включения какой-либо лексики. Согласно узбекскому языку агглютинативности могут быть разработаны конечные автоматы (FSM-конечные автоматы). В отличие от предыдущих работ, в этом исследовании конечные автоматы моделируются для всех классов слов с использованием морфотактических правил в порядке справа налево. В этой статье показаны этапы этой методологии, включая классификацию аффиксов, генерацию конечных автоматов для каждого класса аффиксов и объединение в головную FSM для анализа слова.

**Ключевые слова:** Обработка естественного языка, морфология, словообразование, конечный автомат, удаление аффиксов, узбекская морфология, узбек стемминг.

**Introduction.** Morphological analysis is an important stage on natural language processing. The research in this paper is a contribution to the development morphological analysis tools for Uzbek language. Morphological parsing is the process of dividing a word into its smallest meaningful components called morphemes. In this paper we describe designed Finite State Machine (FSM) based on morphemic rules of the language including related resources in order to do morphological analysis of Uzbek words.

Uzbek language spoken in general in Uzbekistan (and some other places in Central Asia). It is left to right written language. Uzbek language is an agglutinative [10], it is a language whose words are generated by adding affixes to the root forms. In such languages, given a word in its root form we can drive a new word by adding affix, and so on. Thus in many cases, a single Uzbek word may correspond to a many-word sentence or phrase in a non-agglutinative language. The language has a rich morphological structure which is commonly words formed by general structure of morphology (Figure-1) and rarely is not.

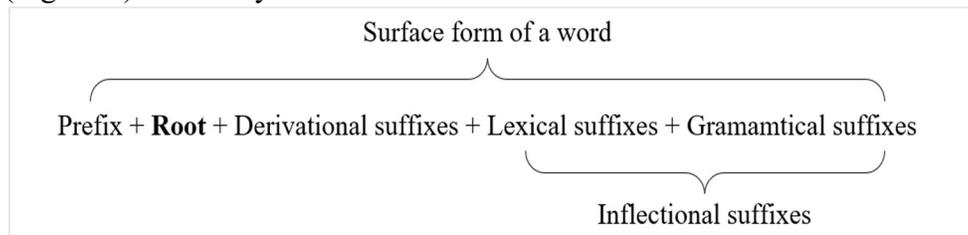

**Figure 1: General morphological structure of Uzbek word**

In Uzbek have prefixes and main part of them are derivational affixes. These have assimilated from other languages. Extracting derivational suffixes from stems are transformed to roots. The derivational suffixes create a new meaning from the root/stem. A word made by derivational affixes belongs to only one of the following word classes: Noun, Adjective, Adverb and Verb.

Inflectional suffixes are divided into two groups, lexical form maker and syntactic form maker, give another point meaning to the word. According suffix content, suffixes grouped into simple suffix and complex suffix. Simple suffix contains only one suffix whereas complex two or more.

In Uzbek, inflectional suffixes generally come after derivational suffixes. Some exceptional derivational suffixes that come after inflectional suffixes are: like o'ch/ir/gich (–gich - derivational suff., –ir - inflectional suff.).

Morphological analyzers are usually constructed using one of the following approaches [Viks, 1994, p7]:

- A dictionary approach, where the main part of the analyzer is a dictionary containing all words' forms in a specific language. Analysis of a given word is reduced to the searching for the matching form in the dictionary, and to reading grammatical information from this dictionary.
- An algorithmic approach, where analyzer include a set of the inflection rules. Related rules are being applied to a given word to extract its base form. Grammatical information depends on the chosen rules.

The first approach ensures high quality of the analysis, but only for the forms which exists in the dictionary. Its main weaknesses include a substantial memory requirements and labour-consuming need for the dictionary development. The most important advantage of the second approach is its ability to analyze forms which do not require a dictionary.

**Literature review.** The research [1] presented the design of a morphological analyzer for Turkish with an affix stripping approach and without using a lexicon. The paper [1] describes the steps of this methodology including the suffixes classification, the construction of the FSMs for each class and all FSMs integrated to a main one which is make a response to morphologic analysis of a given word. While constructing the methodology are based agglutinative structure of the language and the language has only suffix as affix type. In contrast Turkish, the Uzbek language have both affix type: prefix and suffix.

The paper [2] used DCG (Define Clause Grammar) notation to create a Lexicon for morphological analysis of Uzbek language. Because DCG is simplification to encoding lexicon and applied in Prolog language. In the work, suffixes are studied in Derivational and Conjugational classes. They have examined the morphological structure of the Uzbek language, resulting in UZMORPP (Uzbek Morphological Parser) program implemented in Prolog which is including lexicon (1000 entries) and 108 suffixes. While designed the algorithm morphotactic and morphophonemic rules of Uzbek language are included.

In the research [4], the architecture of the morphological analyzer for the Uzbek language is proposed, the functionality is described using the IDEF0 model, and a finite-state transducer for grammatical and morphological rules is created. Here also described a formal mathematical model for rules morphotactic. Based on the rules of grammar, morphotactic, morphonology, an algorithm was developed using a finite state machine and a transducer. For the Uzbek morphological analyzer, an architecture is proposed that works on the basis of the lexicon of words, rules of morphotactic and morphonology. All mechanisms are based on the theory of finite automata.

In the paper [5], study the model of morphological analysis of the Uzbek language, a mathematical model is proposed that takes into account word-forming suffixes, affixes, and other forms of affixes for nouns. Based on the created mathematical model, it developed an algorithm for morphemic parsing and a group of pre-belonging word forms using finite automata, which supports parsing word forms belonging to the lexical category of nouns.

The study [6] includes some decisions on morphological analyzer of the texts on Uzbek to create the antiplaque program. Scientific novelties of this work are: method is proposed for doing the analysis of Uzbek words with an affix stripping approach and without using any lexicon. The rule-based and agglutinative structure of the language allows Uzbek to be modeled with finite state machines (FSMs). steps of this new methodology including the classification of the suffixes, the generation of the FSMs for each suffix class and their unification into a main machine to cooperate in the analysis. Also there has been provided information about dividing affixes into groups and defining the root in the word with the help of FSM.

In contrast from the above works, in this study Uzbek language affixes were classified into 7 classes based on morphotactic constraints and were built FSMs for each class. Consequence the main FSM was built by combining these FSMs. As a result, it was developed an algorithm of the stemmer for Uzbek.

**Research Methodology.** Uzbek, which is an agglutinative language, has a very rich morphological structure. The Uzbek words often contain some semantic information after the multiple affixations. Some of a word construction is not affined to common structure of the word. Ex: The suffix "–*lar*" besides of plural has a another meaning which is indicating his/her greeting to pointed person. Indeed, when the suffix used in a word, it formed by a different morphological structure:

*dada/m/lar*    root(*dada*) + Possessive (*m*) + Greeting (*lar*)       my father
*kitob/lar/im*    root(*kitob*) + Plural (*lar*) + Possessive(*im*)         my books

In general, greeting meaning of the *–lar* suffix comes after possessive and used only for person (the pointed person should be older than talking person). The both affix type, suffixes and prefix are existing in Uzbek. Therefore, when an affix stripping method is used, the analysis is made by removing the prefixes from the beginning of a word as well as the suffixes from the end of the word.

In Uzbek, there are also suffixes that are written in the same form, but have different meanings. Herewith, there are suffixes that are formed by adding two or more suffixes side by side. For example, there are suffixes *–chi*, *–lik*, but there is also a monolithic form suffix *–chilik* according to the rule of separation of morphemes. When analyzing of the suffix *–chilik* can be a combination of 2 suffixes side by side ([*–chi*] +[*–lik*]) or mono form suffix (*–chilik*) depending to the content. There is also a suffix that is contain three suffixes side by side: ([*–gar*]+[*–chi*]+[*–lik*], *odamgarchilik*):
*gul*+[*–chi*]+[*–lik*] (*gulchi* [flower], *gulchilik* [floriculture])
*dehqonchilik* (*dehqonchi*, it is incorrect word)
In the affix stemmer, while the lexicon is not used, it is planned to looking for a longer suffix (an undivided suffix) instead of one of the adjacent suffixes.
With the purpose of composing an affix stripping analyzer for Uzbek, the affixes are firstly classified into classes through their role and coming order. Table-1 shows the generated affix classes.

An affix in Uzbek can have multiple allomorphs in order to provide sound harmony (as the phonological rules) in the word to which it is affixed. For example, the adverb suffix with generic representation *–Gancha* has three allomorphs: *–gancha*, *–kancha*, *–qancha*. The abbreviations used to show suffixes in a generic way are shown below:

**G: g , k, q    Y: a, y    K: k, g    Q: k, g, g', q    T: t, d**
(): the letter between parentheses can be omitted

Table-2 shows the number of affixes and the number of allomorphs for each class.

As the table-2, The number of affixes of all derivational types is 77 and the number of its allomorphs is 87, the number of all affixes of inflectional types is 95 and the number of its allomorphs is 135.

**Table 1: Affix classes**

| Class # | Class name | Type |
|---|---|---|
| 1 | Tense & Person suffixes | Inflectional |
| 2 | Verb suffixes | Inflectional |
| 3 | Relative verb suffixes | Inflectional |
| 4 | Derivational suffixes | Derivational |
| 5 | Noun suffixes | Inflectional |
| 6 | Number suffixes | Inflectional |
| 7 | Prefixes | Derivational |

**Table 2: Count of Affixes and Allomorphs**

| Class # | Affixes | Allomorphs |
|---|---|---|

| 1 | 24 | 27 |
|---|----|----|
| 2 | 23 | 41 |
| 3 | 13 | 23 |
| 4 | 71 | 81 |
| 5 | 23 | 31 |
| 6 | 11 | 12 |
| 7 | 7  | 7  |
| Total | 172 | 222 |

The steps after classified of the affixes is design FSMs for each class which make a reversed order (right to left) analysis of a word. We perform following stages turn by turn to create FSMs:
• designing a left to right FSM;
• Identification the affixes;
• Inverting the left to right FSM and obtaining a non-deterministic finite state automaton (NFA);
• Converting NFA to a deterministic finite automaton (DFA) and constructing the right to left FSM;
In the remaining part of this section, affix table and right to left FSM are given for each affix class. The above four stages are explained in detail on the "Tense & Person suffixes" class since its FSMs and NFA to DFA conversion operations are simpler than the other classes.

**Analysis and results.**
In this parts, it presents the results after generating above stages for each affix classes.

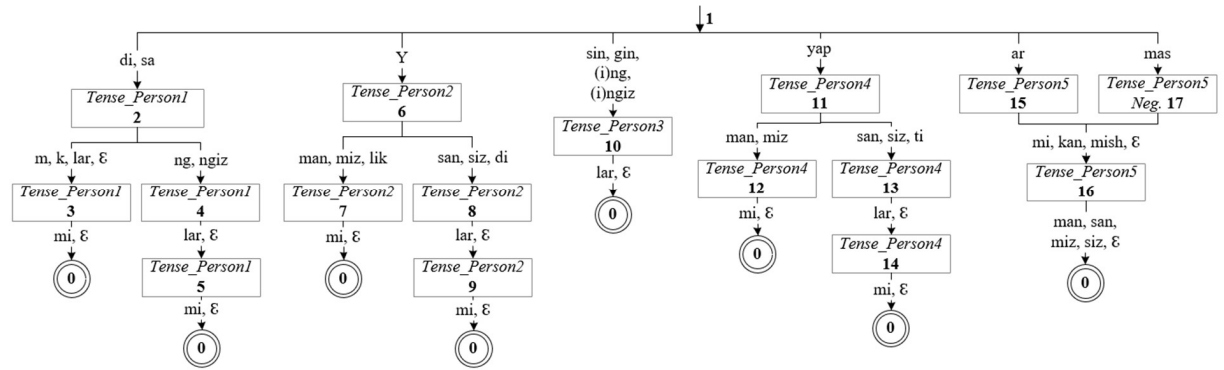

**Figure 2: Tense & Person suffixes left to right FSM**

In Figure-2, a numerical value is indicated a state. In the following stages, the states will be expressed with these numbers: 0 – ending state, 1 – initial state. The character "ε" means the empty transitions between the states. When the analysis of the word "*bor-yap-siz*" (you are going) is made by using this

FSM, firstly the stem "*bor*" (go) is found, then the first suffix "*–yap*" makes a transition from the input state 1 to state 11, after that the suffix "*–siz*" carries the machine to state 13. The ending state is reached by the empty transition between state 13 and 14, also 14 and 0.

Table 3: Tense & Person suffixes

| 1 | –di | 6 | –ng | 11 | –lik | 16 | –(i)ng | 21 | –mas |
|---|---|---|---|---|---|---|---|---|---|
| 2 | –sa | 7 | –ngiz | 12 | –san | 17 | –(i)ngiz | 22 | –mi |
| 3 | –m | 8 | –Y | 13 | –siz | 18 | –yap | 23 | –kan |
| 4 | –k | 9 | –man | 14 | –sin | 19 | –ti | 24 | –mish |
| 5 | –lar | 10 | –miz | 15 | –gin | 20 | –ar | | |

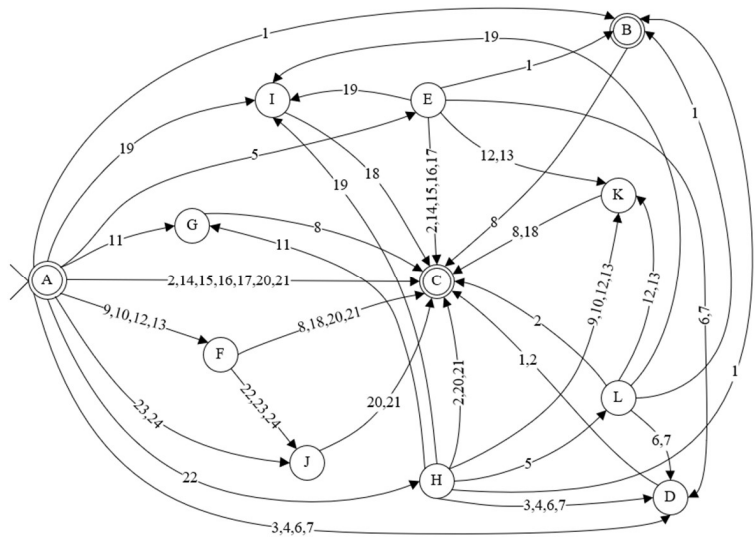

Figure 3: Tense & Person suffixes right to left FSM

Table 4: Verb suffixes

| 1 | –Gan | 6 | –ay | 12 | –gudek | 17 | –(i)sh | 23 | –mayin |
|---|---|---|---|---|---|---|---|---|---|
| 2 | –moqda | 7 | –Ydigan | 13 | –Gani | 19 | –moq | 24 | –masliK |
| 3 | –(a)yotgan | 8 | –(a)yotir | 14 | –(i)b | 20 | –moqliK | 25 | –liK |
| 4 | –moqchi | 10 | –Gancha | 15 | –(u)v | 21 | –ma | | |
| 5 | –lay | 11 | –Guncha | 16 | –(a)r | 22 | –may | | |

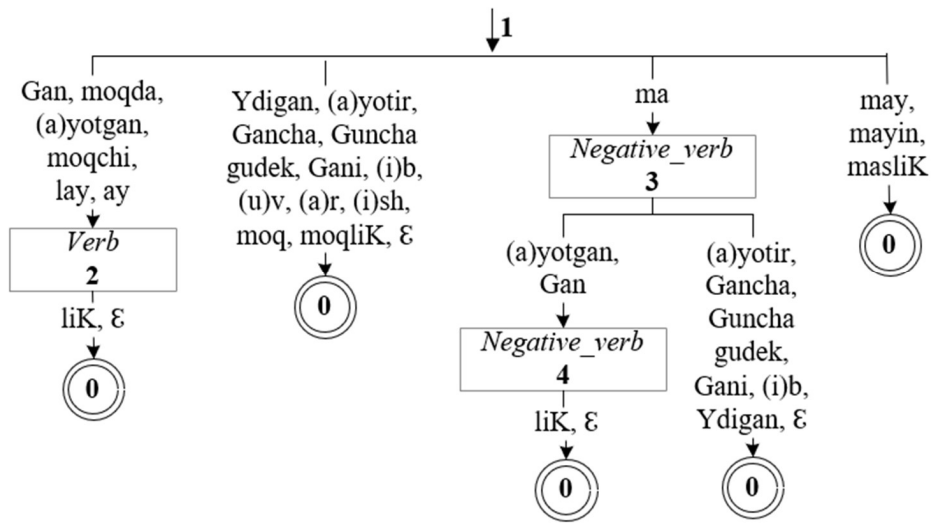

**Figure 4: Verb suffixes left to right FSM**

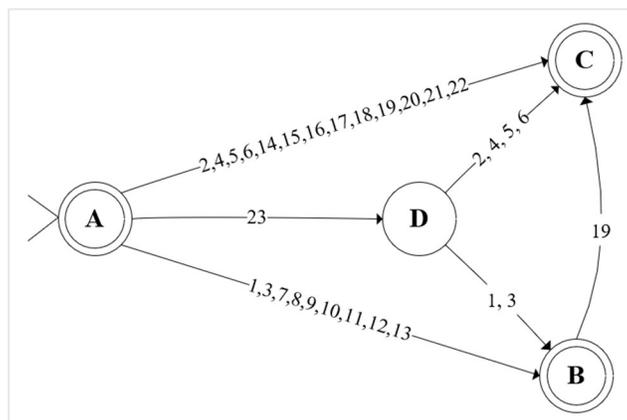

**Figure 5: Verb suffixes right to left FSM**

The relative verb forms are very actively used in Uzbek language while forming a verb. Two or three relative verb suffixes can be continuously added to a verb: *ko'rsattirmoq, undirtirmoq, gapirtirmoq, bezantirildi*. The relative verb suffixes reveal aspects of a lot of meanings of a verb that are hidden in other contexts, affect the transitiveness of a verb, and change the meaning of a verb. It also services to indicate the differences between the morphological-paradigmatic moment, lexical-morphological moment, syntactic moment of words and word forms in the Uzbek language.

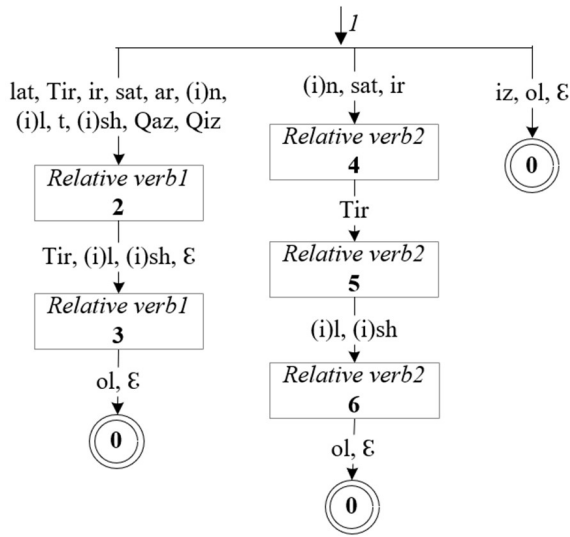

**Figure 6: Relative verb suffixes left to right FSM**

**Table 5: Relative verb suffixes**

| 1 | –lat | 6  | –(i)n  | 11 | –Qiz |
|---|------|----|--------|----|------|
| 2 | –Tir | 7  | –(i)l  | 12 | –iz  |
| 3 | –ir  | 8  | –t     | 13 | –ol  |
| 4 | –sat | 9  | –(i)sh |    |      |
| 5 | –ar  | 10 | –Qaz   |    |      |

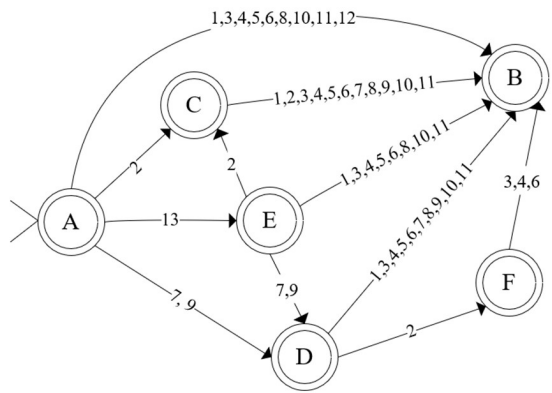

**Figure 7: Relative verb suffixes right to left FSM**

In Uzbek language, there is a way to form a verb consisting of two consecutive words, whereas second one is an auxiliary verb. For example, "*ko'tara olmadi*" ("could not lift"), it is a single verb which is formed using two independent words. In that case, can be written as "*ko'tarolmadi*". This structure made by *–ol* suffix, which is considered an independent word. In such word formation, the –*ol* suffix always comes after the relative verb suffixes. Therefore, this suffix must be placed after the all of relative verb suffixes.

According to the rules of word formation in Uzbek linguistics, we decided to create one FSM for all derivations, because all derivations are attached to the root based on almost the same structure.

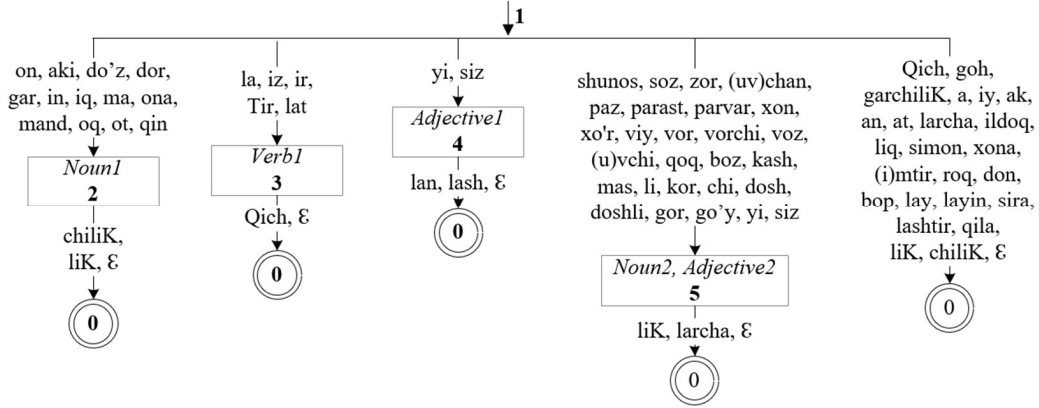

**Figure 8: Derivational suffixes left to right FSM**

**Table 6: Derivational suffixes**

| 1 | –liK | 19 | –Tir | 37 | –vorchi | 55 | –ak |
|---|---|---|---|---|---|---|---|
| 2 | –chiliK | 20 | –lat | 38 | –voz | 56 | –an |
| 3 | –on | 21 | –Qich | 39 | –(u)vchi | 57 | –at |
| 4 | –aki | 22 | –yi | 40 | –qoq | 58 | –larcha |
| 5 | –do'z | 23 | –siz | 41 | –boz | 59 | –ildoq |
| 6 | –dor | 24 | –lan | 42 | –kash | 60 | –liq |
| 7 | –gar | 25 | –lash | 43 | –mas | 61 | –simon |
| 8 | –in | 26 | –shunos | 44 | –li | 62 | –xona |
| 9 | –iq | 27 | –soz | 45 | –kor | 63 | –(i)mtir |
| 10 | –ma | 28 | –zor | 46 | –chi | 64 | –roq |
| 11 | –ona | 29 | (uv)chan | 47 | –dosh | 65 | –don |
| 12 | –mand | 30 | –paz | 48 | –doshli | 66 | –bop |
| 13 | –oq | 31 | –parast | 49 | –gor | 67 | –lay |
| 14 | –ot | 32 | –parvar | 50 | –go'y | 68 | –layin |
| 15 | –qin | 33 | –xon | 51 | –goh | 69 | –sira |
| 16 | –la | 34 | –xo'r | 52 | –garchiliK | 70 | –lashtir |
| 17 | –iz | 35 | –viy | 53 | –a | 71 | –qila |
| 18 | –ir | 36 | –vor | 54 | –iy | | |

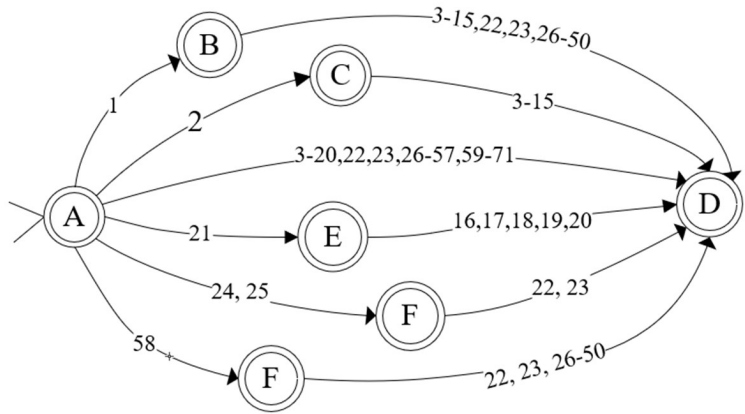

**Figure 9: Derivational suffixes right to left FSM**

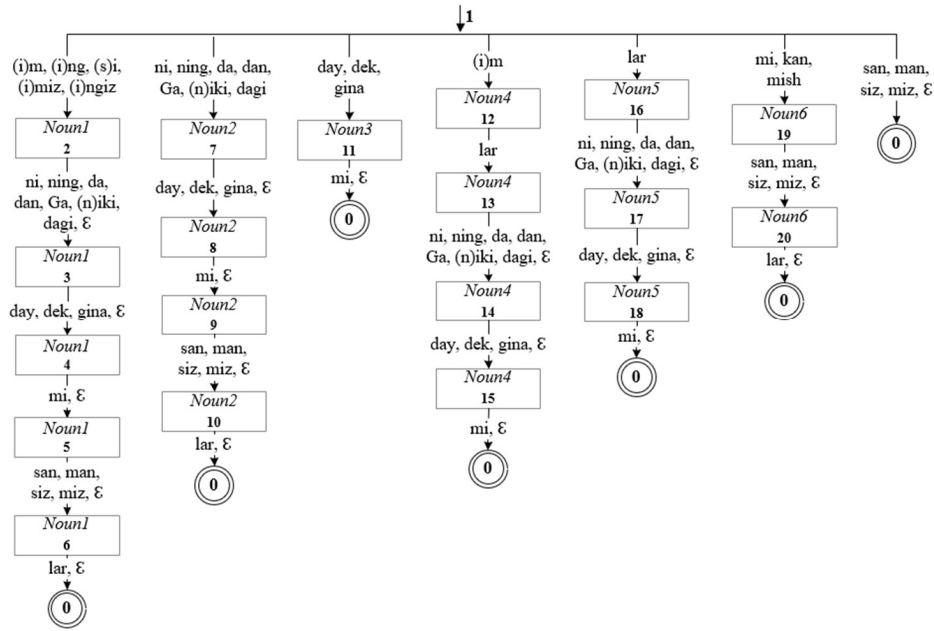

**Figure 10: Noun suffixes left to right FSM**

**Table 7: Noun suffixes**

| | | | | | | | | | |
|---|---|---|---|---|---|---|---|---|---|
| 1 | –(i)m | 6 | –ni | 11 | –(n)iki | 16 | –mi | 21 | –miz |
| 2 | –(i)ng | 7 | –ning | 12 | –dagi | 17 | –lar | 22 | –kan |
| 3 | –(i)miz | 8 | –da | 13 | –day | 18 | –san | 23 | –mish |
| 4 | –(i)ngiz | 9 | –dan | 14 | –dek | 19 | –man | | |
| 5 | –(s)i | 10 | –Ga | 15 | –gina | 20 | –siz | | |

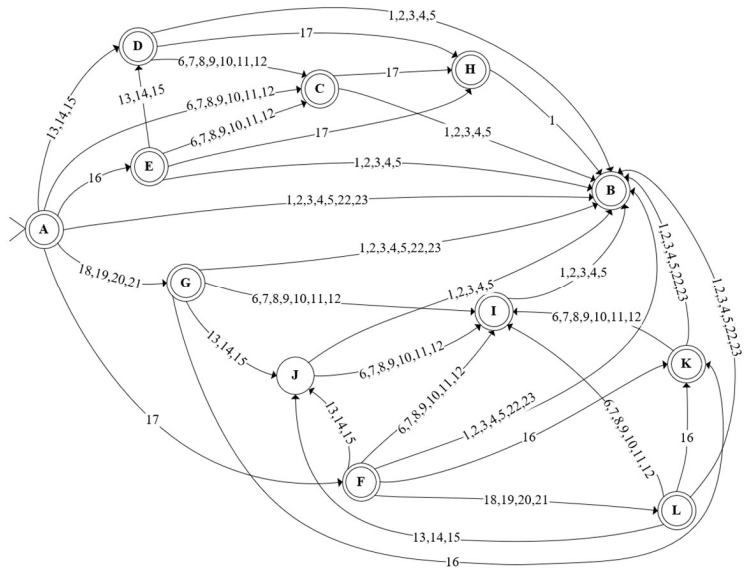

**Figure 11: Noun suffixes right to left FSM**

The number of prefixes in Uzbek is very small. In Uzbek. In general, the prefixes were transformed from Persian language. We conditionally called Prefixes class type is Derivational, because all prefixes are derivational except one *alla–*. *alla–* is inflectional (lexical) suffix. Whether it is a derivational prefix or inflectional prefix, there is only one prefix can be attached before stem, if needed. (As the concatenation rules of Uzbek language, only a single prefix can be occurred in a word.)

Table 8: Prefixes

| 1 | ba– | 4 | bar– | 7 | alla– |
|---|---|---|---|---|---|
| 2 | be– | 5 | no– | | |
| 3 | bo– | 6 | ser– | | |

Table 9: Number suffixes

| 1 | –(i)nchi | 5 | –ala | 9 | –ovlash |
|---|---|---|---|---|---|
| 2 | –ta | 6 | –lab | 10 | –ovlashib |
| 3 | –tacha | 7 | –ov | 11 | –ovlon |
| 4 | –larcha | 8 | –ovlab | | |

In the last step, all created FSMs integrated to the main FSM based on their relationship. The main FSM has two entrance (Tense & Person suffixes and Noun suffixes classes) and two final point (Number suffixes and Prefixes classes).

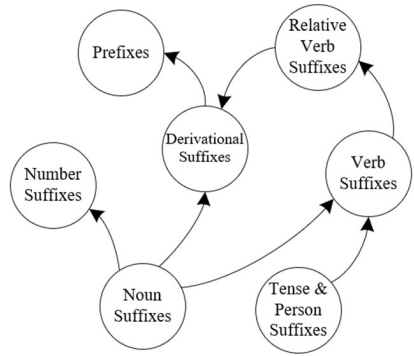

**Figure 12: Relationship of the FSMs**

Example: In the following lines demonstrated the output taken after the analysis of the word "*bajartirilmayaptimi*" ("Is it not being performed?") by the main FSM. In each row include information about of the affix, the definition of the affix and its related affix class:

"*bajar-tir-il-ma-yap-ti-mi*"

| | | |
|---|---|---|
| *bajar* | stem, verb | |
| *–tir* | relative verb suffix | Relative verb suffix |
| *–il* | relative verb suffix | Relative verb suffix |
| *–ma* | negative verb suffix | Verb suffix |
| *–yap* | continuous tense suffix | Tense & Person suffix |
| *–ti* | 3$^{nd}$ single person suffix | Tense & Person suffix |
| *–mi* | question suffix | Tense & Person suffix |

**Conclusion & Future Work**

In this work, an affix stripping morphological analyzer is developed for Uzbek language which is based the morphotactic rules of the language. This model focuses to reach the stem of a word without using any lexicon while making the morphological analysis. To reach this aim, all the affixes are grouped into seven classes. For each of the affix class, the FSM describing the concatenation rules of the suffixes in reverse order is designed. A global FSM is formed to make the previously designed FSMs collaborate with each other. At the end of the analysis with the global FSM, the word is partitioned into its stem and suffixes.

Uzbek language has too more derivational suffixes, even some of them rarely using to make a word. When we want to add new suffixes to this analyzer, we should insert it to the related affix class and update the related FSM.

Morphological analysis is an important component of natural language processing systems like spelling correction tools, parsers, machine translation systems, and dictionary tools.

In Uzbek sentences, words are usually concatenated lots of suffixes, that's reason an information retrieval system needs a quick performing stemming algorithm. Especially in the web-based systems concerning the performance criteria, it is undesirable to use a lexicon to find the word stem.

The considered method in this study can be the basis for the developing an Uzbek stemming algorithm and is being created a complete software for Uzbek stemming according the algorithm. This will be applied to develop a software in information retrieval system for Uzbek documents.